\documentclass{article}

\usepackage{arxiv}

\usepackage[utf8]{inputenc} % allow utf-8 input
\usepackage[T1]{fontenc}    % use 8-bit T1 fonts
\usepackage{hyperref}       % hyperlinks
\usepackage{url}            % simple URL typesetting
\usepackage{booktabs}       % professional-quality tables
\usepackage{amsfonts}       % blackboard math symbols
\usepackage{nicefrac}       % compact symbols for 1/2, etc.
\usepackage{microtype}      % microtypography
\usepackage{lipsum}

\usepackage{framed,multirow}

\usepackage{times}
\usepackage{epsfig}
\usepackage{graphicx}
\usepackage{amsmath}
\usepackage{amssymb}
\usepackage[numbers]{natbib}

\usepackage{makecell}
\usepackage{longtable}

\usepackage{subcaption}
\usepackage[linesnumbered,ruled,vlined]{algorithm2e}
\usepackage{microtype}

\SetKwInput{KwInput}{Input}                % Set the Input
\SetKwInput{KwOutput}{Output}              % set the Output

\title{Chaotic-to-Fine Clustering for Unlabeled Plant Disease Images}

\author{
  Uno~Fang \\
  Deakin-SWU Joint Research Centre on Big Data\\
  Deakin University\\
  Burwood, VIC 3125, Australia \\
  \texttt{uno.fang@deakin.edu.au} \\
   \And
 Jianxin~Li \thanks{Corresponding author: E-mail: jianxin.li@deakin.edu.au} \\
  School of IT\\
  Deakin University\\
  Geelong, VIC 3216, Australia \\
  \texttt{jianxin.li@deakin.edu.au} \\
   \AND
   Xuequan~Lu \\
   School of IT\\
  Deakin University\\
  Geelong, VIC 3216, Australia \\
  \texttt{xuequan.lu@deakin.edu.au} \\
   \And
   Mumtaz~Ali \\
   School of IT\\
  Deakin University\\
  Geelong, VIC 3216, Australia \\
  \texttt{mumtaz.ali@deakin.edu.au} \\
   \And
   Longxiang~Gao \\
   School of IT\\
  Deakin University\\
  Geelong, VIC 3216, Australia \\
  \texttt{longxiang.gao@deakin.edu.au} \\
  \And
   Yong~Xiang \\
   School of IT\\
  Deakin University\\
  Geelong, VIC 3216, Australia \\
  \texttt{yong.xiang@deakin.edu.au} \\
}

\begin{document}
\maketitle

\begin{abstract}
Current annotation for plant disease images depends on manual sorting and handcrafted features by agricultural experts, which is time-consuming and labour-intensive. In this paper, we propose a self-supervised clustering framework for grouping plant disease images based on the vulnerability of Kernel K-means. The main idea is to establish a cross iterative under-clustering algorithm based on Kernel K-means to produce the pseudo-labeled training set and a chaotic cluster to be further classified by a deep learning module. In order to verify the effectiveness of our proposed framework, we conduct extensive experiments on three different plant disease datatsets with five plants and 17 plant diseases. The experimental results show the high superiority of our method to do image-based plant disease classification over balanced and unbalanced datasets by comparing with five state-of-the-art existing works in terms of different metrics. 
\end{abstract}

\keywords{Under-clustering \and Similarity Learning, Self-supervised Learning \and Automated Annotation}

%\linenumbers

%% main text
\section{Introduction}

Plant disease detection, one of the most popular topics in Precision Agriculture (PA), has attracted remarkable attentions recently. A general procedure of plant disease detection system consists of five fundamental steps: image acquisition, dataset annotation, image processing, feature extraction, classification \citep{shruthi2019review}. To our knowledge, the majority of studies of plant disease detection studied on improving image processing, feature extraction and image classification based on the well-labelled public datasets. For example, \cite{brahimi2018deep}, \cite{zhang2018identification}, \cite{hu2019low} and \cite{fuentes2018high} introduced novel convolutional neural networks (CNNs) as image classifiers and/or proposed data segmentation and augmentation methods to increase the recognition and classification accuracy. By contrast, the dataset annotation, as a fundamental basis for other processing in plant disease detection, has been sparsely handled. Data annotation is often being conducted manually by agricultural experts, which consumes a lot of labour, material resources and time. Therefore, it expects a unsupervised learning method to enhance the efficiency and effectiveness of plant disease dataset annotation to minimise the manual cost from agricultural experts.

The essence of grouping plant disease images is to extract decent representations of input images and to develop metrics to measure the pairwise similarity for categorisation. Supervised learning can hardly deal with this problem, as it greatly depends on the labeled data and needs to learn all types of data to achieve decent classification outcomes. Therefore, the dataset annotation robustly relies on unsupervised learning and self-supervised learning (clustering). Previous clustering algorithms, such as K-means, Kernel K-means and Spectral clustering, can be utilised to cluster unlabeled plant disease images. However, these algorithms tend to work for balanced data, and greatly rely on the prior knowledge (e.g. ground-truth class number etc). Although there are rarely few clustering methods on plant disease dataset annotation, many state-of-the-art unsupervised methods were proposed for unlabeled face images, such as Approximate Rank Order Clustering (AROC) \citep{otto2017clustering} and FINCH \citep{sarfraz2019efficient}, both of which are graph-based clustering. These methods are developed to deal with various application scenarios from social media to law enforcement, where the number of faces in the collection can reach hundreds of millions. AROC is designed to cluster large-scale unbalanced data since it is developed based on rank order clustering which is often utilised to handle imbalanced data. However, it still faces difficulties in dealing with balanced data. In addition, FINCH is a ``parameter-free'' clustering algorithm, which, on the other hand, somewhat loses certain control during clustering. It essentially utilises the first neighbour of each point to produce the clustering hierarchy and iteratively merges clusters. In this work, a plant has different diseases which appear to be similar in images, such as potato early blight and potato late blight, if such images are clustered/merged together by FINCH, it would pose negative impact on the clustering accuracy. As for the self-supervised learning, \cite{yang2016joint}, \cite{jiang2016variational}, \cite{guo2017improved} and \cite{caron2018deep} combined clustering methods (mostly K-means) and deep learning models where the clustering results are attached with pseudo-labels to train deep learning models, but they greatly depend on priori knowledge and the number of categories of the input dataset. They emphasise over-clustering principally as a reliable approach to allow CNNs to learn representations, but over-clustering would lead to the loss of focus and poor templates on representations. 

To address the grouping of plant disease image datasets without knowing the data structure, we propose a novel framework in this paper. Our core observation is that Kernel K-means can easily produce some accurate clusters and one chaotic cluster (i.e., abnormal cluster) with tuning the threshold. Motivated by this, we attempt to first produce some accurate clusters and an abnormal cluster with a novel cross iterative clustering algorithm. These clusters are then merged, and fed into a simple image classifier for Top-$\mu$ results and a dual-component similarity measurements module including a ResNet50 \citep{he2016deep} based similarity finder and a Siamese similarity network. The final result is computed as the average of the two similarity scores in the collaborative module. Our approach is denoted as CIKICS (\textbf{C}ross \textbf{I}terative \textbf{K}ernel K-means enhanced with \textbf{I}mage \textbf{C}lassification and \textbf{S}imilarity Measurements).  We perform exhaustive evaluation on the proposed CIKICS by conducting comparative experiments to demonstrate its outstanding performance comparing to existing methods on image clustering, in which CIKICS has great potential to advance the existing system of plant disease detection.
The contributions of this paper are as follows.

\begin{itemize}
    \item We develop a under-clustering algorithm (cross iterative Kernel K-means), by leveraging the sensitivity of the parameter in the Radial Basis Function (RBF) Kernel. It produces many small but highly accurate clusters, which are merged and pseudo-labeled as the training set for follow-up deep learning module.

    \item To classify the non-clustered data, we introduce the collaboration of a CNN-based image classifier and two similarity measurements (i.e., a pre-trained CNN-based similarity finder and a trained Siamese network).
    
     \item We design the per-cluster quality measure to evaluate the performance of clustering algorithms, as a complement of existing measures.
    
    \item We conduct extensive experiments on three differently structured datasets to verify the performance of the proposed clustering framework using a set of metrics.

\end{itemize}

The remaining of this paper is structured as follows. In Section 2, we review the related work. Section 3 elaborates our proposed framework. Experimental settings and results analysis are presented in Section 4. Lastly, Section 5 provides the conclusion.

\section{Related Work}

\subsection{Unsupervised Clustering Algorithms}

\subsubsection{\textbf{Partitioning Clustering}}
For center-based clustering, K-means is a widely used but fragile algorithm which is sensitive to the selection of the initial K centroids. Fortunately, K-means++ \citep{arthur2006k} was proposed to address the initial centroids selection issue. In addition, instead of using Euclidean distance, Kernel K-means utilises kernel to calculate the distance for clustering, which can cluster non-linearly separable data \citep{dhillon2004unified}. A Kernel function utilises a decent nonlinear mapping from the original space to a higher-dimensional feature space. If we directly adopt a Kernel function to the original data, it would lead to a high cost of time and the low accuracy, from our observation. As \cite{santhi2018performance} illustrated, three assumptions were expected to be fulfilled so that the use of K-means (including K-means++) would succeed in researching, where (i) the variance of the distribution of each attribute (variable) is assumed to be spherical in K-means; (ii) the variances of all variables are the same and (iii) and observations are clustered into each cluster approximately equally. Additionally, K-medoid (such as PAM, CLARA and CLARANS) \citep{schubert2019faster} has similar issues with K-means \citep{kaufman1990partitioning}. Therefore, to reach these three assumptions would restraint the applicability of these algorithms on different datasets. Most of spectral clustering algorithms \citep{ng2002spectral} need to calculate the complete similarity graph Laplacian matrix and even have secondary complexity, which severely limits their scalability to large data sets. Same as Kernel K-means, spectral clustering is used to obatin non-linear separable clusters in the input space. Although their performances are comparable, there is limitation of these algorithms, where they tend to group the data evenly rather than clustering it according to similarity for matching the ground truth.

\subsubsection{\textbf{Agglomerative/Divisive Hierarchical Clustering}}

Density-based clustering algorithms, such as DBSCAN, OPTICS and DENCLUE, have the capability of clustering non-spherical shaped clusters \citep{chauhan2014survey}. Although DBSCAN is the basis of all density-based clustering algorithms, it is sensitive to outliers and inflexible to varying density data since it is not developed to cluster every data point \citep{chauhan2014survey}. To achieve our research goal, all data points need to be clustered. However, these algorithms cannot be applied. There are many Hierarchical Agglomerative Clustering (HAC) algorithms \citep{zhu2011rank,lin2017proximity,lin2018deep}, but these algorithms usually have the weak capability of clustering large-scale data due to the computation complexity. To deal with large-scale clustering efficiently, AROC was proposed by \cite{otto2017clustering}, which is a method based on rank order clustering with Approximate Nearest Neighbour (ANN) targeting to estimate whether there shall be an edge between a node and its $k$ Nearest Neighbours ($k$NN). \cite{sarfraz2019efficient} introduced a parameter-free clustering algorithm, FINCH, on the basis of the hierarchical clustering algorithm, adopting the notion of the first neighbour relationship. The above mentioned algorithms have been evaluated with good performance on unconstrained face images grouping. However, these methods are not transferable as they are application oriented. Most importantly, some of them \citep{lin2018deep,otto2017clustering} were density-aware and were devised to process data with unbalanced density.

\subsection{Learning-based Self-supervised Clustering Methods}

Recently, besides the above state-of-the-art unsupervised clustering algorithms, there is an increasing number of self-supervised clustering methods based on deep learning. For example, using an auto-encoder to optimise and learn features with existing clustering methods \citep{yang2016joint,guo2017improved,tian2017deepcluster,caron2018deep}. \cite{yang2016joint} proposed a recurrent framework which contains a forward pass of agglomerative clustering and a backward pass of CNN-based representation learning. \cite{guo2017improved} proposed an Improved Deep Embedded Clustering (IDEC) algorithm to address the problem of data structure preservation. These deep learning-based methods mainly use existing clustering methods as a feature representation learning scheme for training deep learning models \citep{yang2016joint,jiang2016variational,guo2017improved,tian2017deepcluster} or as a means of generating pseudo-labels \citep{caron2018deep}. However, These methods have a higher potential to fail when clusters with similar features exist, which limits their application scenarios, especially when dealing with plant disease images, i.e. a plant has several diseases, and different lesions may not be identified correctly. Moreover, it is brilliant to include pseudo-labels in the self-supervised framework \citep{caron2018deep}, but the framework design is not effective for high-accurate clusters.

%\begingroup
\begin{figure*}[ht]
\centering
\includegraphics[width=0.88\textwidth]{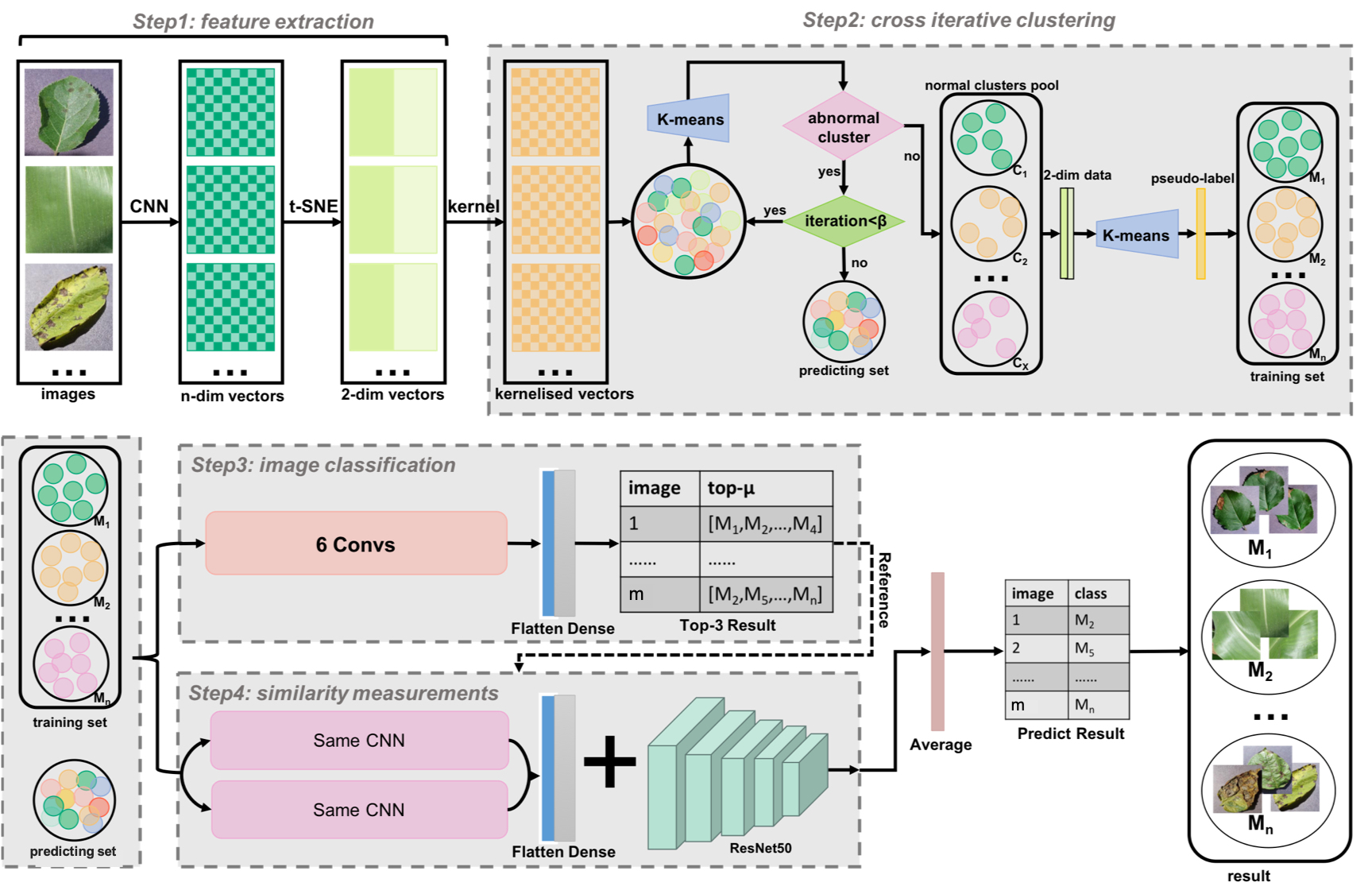}
\caption{Overview of the proposed CIKICS.}
\label{fig:figure3}
\end{figure*}

\section{Overview of Proposed Approach}

As demonstrated in Figure \ref{fig:figure3}, the framework consists of four functional parts, which are feature extraction, cross iterative clustering, image classification and similarity measurements. In the feature extraction step, the n-dimensional representations are extracted by leveraging the ResNet50, and are reduced into 2-dimensional vectors with t-SNE \citep{maaten2008visualizing}. In the cross iterative clustering step, the 2-dimensional features are first transformed into kernelised vectors which can easily enable a chaotic abnormal cluster and some small-sized highly accurate normal clusters. These clusters are merged and pseudo-labeled as the training set. In step 3 and Step 4, we introduce the collaboration of a CNN-based image classifier, a ResNet50-based similarity finder and a Siamese network, which are pre-trained or trained over the above obtained training set to predict the groups of the images in the abnormal cluster. 

\subsection{Plant Disease Image Representation}
\label{subsec:plant_disease_representation}

% \subsubsection{\textbf{Pre-trained CNN architectures}}

We utilise pre-trained CNN architectures, ResNet50, VGG16, VGG19 \citep{simonyan2014very}, Xception \citep{chollet2017xception}, Inception V3 \citep{szegedy2016rethinking} and MobileNet \citep{howard2017mobilenets} as options, to extract image features and to calculate the Euclidean distance representing similarity for each pair of images in the abnormal cluster. We compare these CNN architectures, and utilise K-means as the baseline to evaluate their performance on the dataset PVAG (Table \ref{tab:datasets_details}). Their performances are shown in Table \ref{tab:pre_cnn_precision_table}, where we can obviously observe a prominent capability of ResNet50. Thus we adopt ResNet50 in our experiments.

\begin{table}[h]
    \centering
    \caption{Performance of pre-trained CNN Architectures.}
    \begin{center}
\begin{scriptsize}
 \begin{tabular}{||p{0.21\textwidth}|p{0.06\textwidth}|p{0.06\textwidth}|p{0.06\textwidth}|p{0.06\textwidth}|p{0.06\textwidth}|p{0.055\textwidth}||} 
 \hline
 CNN Model & PRE & Rec & F1 & NMI & PUR & Run (sec) \\ [0.5ex] 
 \hline\hline
 ResNet50 \citep{he2016deep} & \textbf{0.958} & \textbf{0.214} & \textbf{0.349} & \textbf{0.272} & \textbf{0.396} & \textbf{171} \\
 \hline
 Xception \citep{chollet2017xception} & 0.460 & 0.099 & 0.163 & 0.088 & 0.313 & 462 \\
 \hline
 InceptionV3 \citep{szegedy2017inception} & 0.349 & 0.075 & 0.123 & 0.040 & 0.267 & 547 \\
 \hline
 MobileNet \citep{howard2017mobilenets} & 0.811 & 0.177 & 0.290 & 0.248 & 0.396 & 283 \\
 \hline
 VGG16 \citep{simonyan2014very} & 0.896 & 0.198 & 0.324 & 0.259 & 0.375 & 174 \\
 \hline
 VGG19 \citep{simonyan2014very} & 0.916 & 0.204 & 0.334 & 0.267 & 0.377 & 183 \\
 \hline
\end{tabular}
\end{scriptsize}
\end{center} 
    \label{tab:pre_cnn_precision_table}
\end{table}

% \subsubsection{\textbf{t-Distributed Stochastic Neighbor Embedding (t-SNE)}}

Briefly in our case, denote a feature vector set by $X = \{x_{i},\ldots,x_{N}\},x_{i}\in \mathbb{R}^{512}$, each feature vector is extracted from each image in the datatset by ResNet50. %Then we decrease the vector dimensionality to improve computation efficiency using t-SNE.} 
Then we use t-SNE to calculate the similarity probability $p_{ij}$ of point $x_{i}$ and $x_{j}$ in the feature vector set $X$ in the $512$-dimensional space, and map the similarity probability $q_{ij}$ of point $x'_{i}$ and $x'_{j}$ in the corresponding feature vector $X'$ ($X' =\{x'_{i},\ldots,x'_{N}\}, (x'_{i}\in \mathbb{R}^{2})$) in the $2$-dimensional space. This also allows a dimensionality reduction.

\subsection{Cross Iterative Kernel K-means Clustering}

Given that we need to cluster unlabeled data into multiple classes, clustering algorithms are essential in this work. 
Considering that the clustering data is non-linearly separable, we adopt the Kernel K-means \citep{dhillon2004unified}, which was developed based on K-means clustering algorithm with leveraging a kernel function. 

\textbf{Radial Basis Function (RBF) Kernel.}
The kernelisation targets to increase the dimensionality of vectors by comparing vectors pairwise. This way facilitates higher clustering accuracy, according to our empirical observation. We select the RBF kernel \citep{vert2004primer} which is a widely used kernel function in diverse kernelised learning algorithms. Given that we have obtained the feature vector set $X'$ from t-SNE, the RBF kernel \citep{vert2004primer} on two points \(x'_{i}\) and \(x'_{j}\), demonstrated as feature vectors in higher dimensional input space, is defined as Eq. \ref{eq:kernel_1}.
\begin{equation}
\label{eq:kernel_1}
    K(x'_{i},x'_{j}) = \exp \left (-\frac{\left \| x'_{i}-x'_{j} \right \|^{2}}{2\sigma ^{2}}  \right ),
\end{equation}
where $K(x'_{i},x'_{j}) = K(x'_{j},x'_{i})$, and $\left \| x'_{i}-x'_{j} \right \|^{2}$ refers to the squared Euclidean distance between two feature vectors. Due to the fact that the value of the RBF kernel reduces with increasing distance, and it varies between 0 (in the limit) and 1 (when $x'_{i} = x'_{j}$). The kernelised feature vector induce a symmetric matrix in $N$-dimensional space as $Z$ ($Z = \{z_{1},z_{2},\ldots,z_{N}\}, (z_{i}\in \mathbb{R}^{N})$).

\textbf{K-means++ Algorithm.} 
In the framework, we apply K-means++ algorithm \citep{arthur2006k} instead of standard K-means. K-means++ can ensure more intelligent initialisation of the centroids and improve the quality of clustering. Given $Z$ ($Z = \{z_{1},z_{2},\ldots,z_{N}\}, (z_{i}\in \mathbb{R}^{N})$) obtained from the RBF kernel function, we define an integer $k$ (the value of $k$ can be defined around $2\sqrt[2]{N}$) for K-means++ to choose $k$ centers $C$. $D(z)$ represents the shortest distance from a data point to the closest center in $C$. 
%The procedure of K-means++ clustering is shown in Algorithm \ref{al:kmeans_al}.
% \iffalse
% \begin{algorithm}[ht]
% \SetAlgoLined
% \KwInput{Data points $Z = \{z_{1},z_{2},\ldots,z_{N}\}$, Pre-defined cluster number $k$}
% \KwOutput{Clusters $\mathbf{C} = \{C_{1}, C_{2}, \ldots, C_{k}\}$}

% \While{$C \neq \{c_{1},c_{2},\ldots,c_{k}\}$}{
% Select one center $c_{1}$ arbitrarily from $Z$.\;
% Select a new center $c_{i}$, choosing $z \in Z$ with probability $\displaystyle\frac{D(x)^{2}}{\sum_{z\in Z}D(x)^{2}}$.\;
% }

% \While{$\mathbf{C}$ changes}{
% \For{$i = 1,2,\ldots,k$}{
% Set the cluster $C_{i}$ contain points in $Z$ that are closer to $c_{i}$ than other centers\;
% }
% \For{$i = 1,2,\ldots,k$}{
% set $c_{i}$ to be the center of group of all points in $C_{i}$\;
% }
% }

%  \caption{K-means++ Clustering}
%  \label{al:kmeans_al}
% \end{algorithm}
% \fi

\textbf{Cross Iterative Clustering Algorithm.}
As known, $\sigma$ in Eq. \ref{eq:kernel_1} is a parameter that is set prior to the kernel computation. $\sigma$ is utilised to configure the sensitivity to differences in feature vectors, which conversely relies on various characteristics, such as input space dimensionality, data size and feature normalisation. According to Eq. \ref{eq:kernel_1}, we can obtain:

\begin{equation}
    \sigma = \frac{-\left \| x'_{i}-x'_{j} \right \|}{\sqrt{2}\ln{K(x'_{i}, x'_{j})}}
\end{equation}

\paragraph{Breakthrough}
In the case when $\sigma \rightarrow 0$, $K(x'_{i}, x'_{j}) \rightarrow \infty$, which means even similar elements will be pushed apart in the kernel space. Otherwise, when $\sigma \rightarrow \infty$, $K(x'_{i}, x'_{j}) \rightarrow 0$, which means even dissimilar elements will be gathered in the same cluster. We empirically found that appropriately considering more attributes of the applied dataset as variables in manipulating the value reduction of $\sigma$ could lead to a higher accuracy of clusters. However, one of these clusters (the abnormal cluster), which contains kernelising-pushed-away elements, is chaotic and of dramatically large size. \textbf{This is our key observation and the foundation of our framework}, which will be demonstrated in details below. This abnormal cluster enables us to build the cross iterative clustering algorithm, which will filter out images that are hardest to cluster.

\begin{figure*}[ht]
\centering
\includegraphics[width=.68\textwidth]{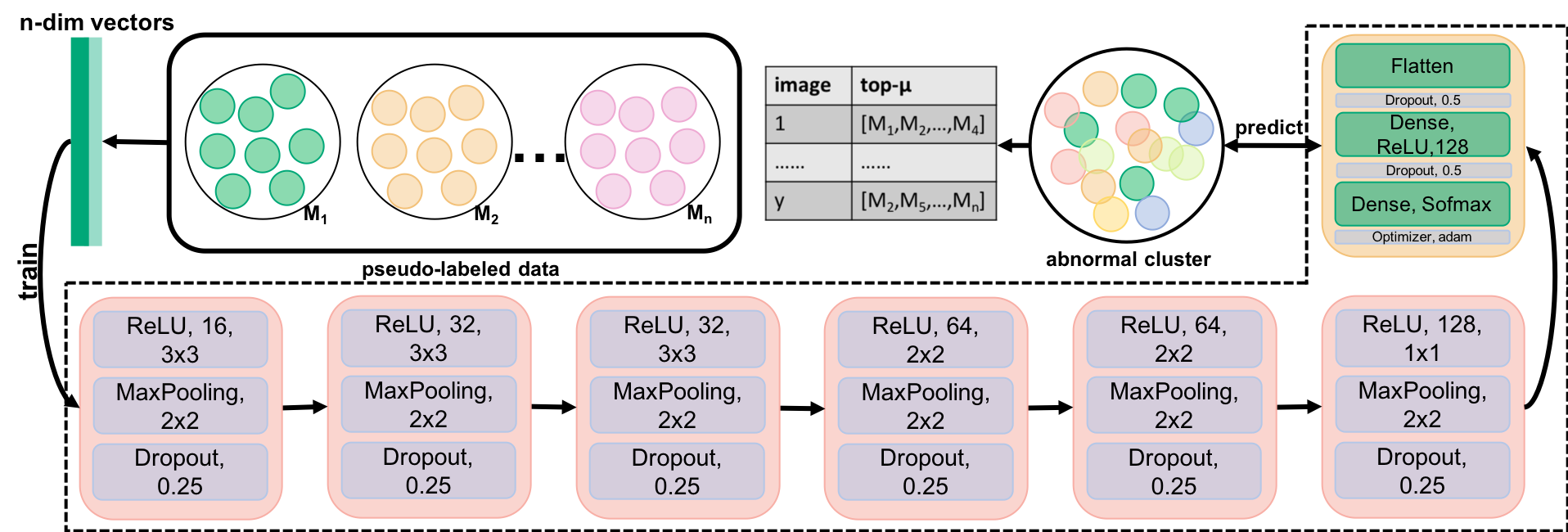}
\caption{Structure of image classification based on CNN. }
\label{fig:image_classification}
\end{figure*}

\paragraph{Parameters}
To obtain the ``optimal'' value of $\sigma$ to kernelise time series data, we refer to a Global Alignment Kernel (GAK) method proposed by \cite{cuturi2011fast}. The bandwidth $\sigma$ is set as a multiple of the median of Euclidean distance of pairwise points observed in different time series of the training set, scaled by the square root of the maximum number of time points for each data point (the input space data dimensionality) $N$. It is defined as:

\begin{equation}
\label{eq:gak}
    \sigma = median(\left \| x'_{i}-x'_{j} \right \|^{2}) \times \sqrt{N}
\end{equation}

As shown in Eq. \ref{eq:gak}, only data points and their space dimensionality are considered as parameters. To establish the cross iterative clustering, we need to involve additional parameters in defining $\sigma$. There are few parameters to define, which are the iteration time $\beta$, the data length $N$ and the pre-defined cluster number $k$. We re-define their relations based on Eq. \ref{eq:gak} as:

\begin{equation}
\label{eq:sigma}
    \sigma = median(\left \| x'_{i}-x'_{j} \right \|^{2}) \times \frac{\ln{k}}{\ln{N} \times \beta}
\end{equation}

We exclude the factor of the input space data dimensionality $d$ because it is an invariant in our framework.

\paragraph{Algorithm}
Given above, we develop the Cross Iterative Kernel K-means as shown in Algorithm \ref{al:cross_iterative_clustering}, and we expect to annotate the merged normal clusters $M$ with pseudo-labels, e.g. first cluster will be labeled as ``M$_{1}$'', second as ``M$_{2}$'', etc. The abnormal cluster and incorporated normal clusters with pseudo-labels produced by the Cross Iterative Kernel K-means provide the flexibility to the clustering framework for deep learning models.

\begin{algorithm}[ht]
\SetAlgoLined
\KwInput{t-SNE feature vector $X'$, RBF Kernel $K(\cdot)$, K-means++ $F(\cdot)$}, iteration frequency $\beta$, coalescence number $n$
\KwOutput{merged normal clusters $M$, abnormal cluster $A$}
$K(X')=Z=\{z_{1},z_{2},\ldots,z_{N}\}$\;
Build an empty array $G_{nm}$ for collecting normal clusters\;
Set the initial abnormal cluster $A=Z$\;
 \For{$i=1,2,\ldots,\beta$}{
 $F(A)=\mathbf{\mathbf{C}}=\{C_{1},C_{2},...,C_{k}\}$\;
 \For{$j=1,2,\ldots,k$}{
 Find the cluster with the max size as the abnormal cluster $C_{j}$\;
 Set $A=C_{j}$\;
 }
 Insert the rest $k-1$ normal clusters into $G_{nm}$\;
 }
 Obtain the final $A$\;
 Obtain $G_{nm}=\{G_{1},G_{2},\ldots,G_{x}\},h = (k-1) \times \beta$\;
%  Set $x= (k-1) \times \beta$\;
 Set an new array $B$ for collecting the centroid of points in each normal cluster of $G_{nm}$\;
 \For{$i=1,2,\ldots,h$}{
 Get all point data of $G_{i}$ from $X'$\;
 Calculate the centroid $a_{i}$ of $G_{i}$\;
 Insert $a_{i}$ into $B$;
 }
 Obtain the merged clusters $M = F(B)$
 \caption{Cross Iterative Clustering}
 \label{al:cross_iterative_clustering}
\end{algorithm}

\subsection{Image Classification with CNN}

Given pseudo-labeled normal clusters $M$ as the training set, and the final abnormal cluster $A$ as predicting set, we involve CNN-based image classifier to learn features of images in the training set $M$, then predict the top-$\mu$ classes of each image $x_{i}$ in $A$. The detailed deployment of our image classifier is demonstrated in Figure \ref{fig:image_classification}. We conduct a comparative experiment on the dataset PVAG  (Table \ref{tab:datasets_details}) with different $\mu$ in the range $(1,n)$, where $n$ is the number of clusters in the training set, to compare the accuracy of the correct classes in the top-$\mu$ list, and the runtime for further similarity measurements. We observe that the accuracy is $0.97\pm0.02$ in the range, but the runtime differs greatly since more potential classes lead to higher computation complexity for deep learning models. Additionally, we observe that, when $\mu=3$, the framework works efficiently and effectively.

\begin{table}[ht]
    \centering
    \caption{Different Top-$X$ Potential Class List Comparison. }
    \begin{center}
\begin{scriptsize}
 \begin{tabular}{||p{0.16\textwidth}|p{0.07\textwidth}|p{0.07\textwidth}|p{0.07\textwidth}|p{0.07\textwidth}|p{0.07\textwidth}|p{0.07\textwidth}||} 
 \hline
  & Top-1 & Top-2 & Top-3 & Top-4 & Top-5 & Top-6 \\ [0.5ex] 
 \hline\hline
 Accuracy & 0.964 & 0.965 & 0.976 & 0.976 & 0.988 & 0.988 \\
 \hline
 Run-time (m) & 5.75 & 8.37 & 9.55 & 13.14 & 16.38 & 21.70 \\
 \hline
\end{tabular}
\end{scriptsize}
\end{center} 
    \label{tab:top_3_evaluation}
\end{table}

\subsection{Similarity Measurements}

Although a simple image classifier is reliable to detect edges, textures and colours, it is not proficient in obtaining high-level features, such as lesion features. Also, the noise in the training set might have negative impact on the image classifier. As such, we include two more similarity measurements as weights to help classifying images in the abnormal cluster into the merged normal clusters $M$.

\iffalse
\begin{figure*}[ht]
\centering
\includegraphics[width=0.82\textwidth]{figures/siamese_network.png}
\caption{Siamese Network}
\label{fig:siamese_network}
\end{figure*}
\fi

\subsubsection{\textbf{Similarity Finder with Pre-trained CNN}}

First of all, we utilise ResNet50 to calculate the similarity score of each image in the abnormal cluster to classes of the training set (normal clusters with pseudo-labels), by comparing the Euclidean distance. In particular, the similarity of two image feature vectors $x_{i}$ and $x_{j}$ is defined as their squared Euclidean distance \citep{wang2014learning}:
\begin{equation}
    D(x_{i}, x_{j}) = \left \| x_{i} - x_{j} \right \|_{2}^{2}
\end{equation}
The smaller the distance $D(x_{i}, x_{j})$ is, the more similar the two images $x_{i}$ and $x_{j}$ are. We cast the similarity ranking problem as a neighbour search problem in Euclidean space. From this similarity finder, for each predicting image, we can get its similarity scores $S(x_{i})$ ($x_{i} \in A$) towards the top-$\mu$ classes (i.e. pseudo labels) from $M$. 

\subsubsection{\textbf{Siamese Network}}

We introduce a Siamese network as a second similarity measure to calculate the similarity scores. It can be finally decided what classes those images in the abnormal cluster shall be classifies into. Similar to the image classifier (above section), we train the Siamese network on the training set, and utilise the trained model to calculate the similarity score of each image in the abnormal cluster to each class in the top-$\mu$ classes. Given the nature of Siamese network, we input two same-sized groups of images to compare the similarity between them.

In the Siamese network, images are compared in paired groups, the \textit{sigmoid} layer will output the similarity score. We have obtained a predicting image set $A$ (i.e. the final abnormal cluster) and a pseudo-labeled image set $M$, where $M = \{M_{1},\dots,M_{n}\}$ and $n$ indicates the number of pseudo-labeled clusters. We utilise Siamese Network to compute the similarity score between $x_{i}$ in $A$ and $M_{j}$ (a pseudo-labeled cluster) in $M$. For computation purpose, we need to expand $x_{i}$ (a feature vector of an image) to $A'_{i}$ (a feature vector set) with blank images, where the size of $A'_{i}$ is expected to be equal to the size of $M_{j}$. The similarity score between $A'_{i}$ and $M_{j}$ is $S'(x_{i}) = \displaystyle\frac{\sum_{i=1}^{n} s'_{i}}{n}$.

% Suppose we have two groups of images \(P\) and \(Q\), and \(P_{i} \in P\), \(Q_{i} \in Q\), the similarity score between \(P\) and \(Q\) is \(S_{i} = \displaystyle\frac{\sum_{i=1}^{n} s_{i}}{n}\), where \(n\) is the number of pairs, and $s_{i}$ ($s_{i} = (P_{i},Q_{i},Y)$) is the similarity score of each paired images.

% The loss function $l(P_{i},Q_{i},x'_{i})$ for a pair with similarity label \(Y\) could be defined as:

% \begin{equation}
% \begin{split}
%     l(P_{i},Q_{i},x'_{i})= x'_{i}\cdot D(f(P_{i}),f(Q_{i})) \\
%     + (1 - x'_{i})\max (0,C-D(f(P_{i}),f(Q_{i}))), \\
%     \forall I_{i},P_{i},Q_{i}\in \left \{ 0,1 \right \},
% \end{split}
% \end{equation}
% where \(C\) is a margin value. This is a margin-based contrastive loss function, which allows to learn a margin of separation. If clusters are similar, the value of \(Y\) is 1; otherwise, it is 0. 

\textbf{Final class determination.} Through the Siamese network, we get a second similarity score $S'(x_{i})$ for $x_{i}$ in $A$. Finally, we calculate the average of $S'(x_{i})$ and $S(x_{i})$ (i.e. the similarity score obtained in Similarity Finder), and the class with the highest average score will be the image's final class.

\section{Experimental Results}
\label{sec:results}
\subsection{Dataset and Preprocessing}
%In many recent researches on plant diseases, 
In this work, we use a well-labeled plant disease image dataset, PlantVillage \citep{hughes2015open}, which contains 54,305 images from 14 kinds of plants and 38 plant diseases, for our experiments. Since unlabeled datasets are expected in our work, we de-label the PlantVillage dataset. Furthermore, to fit the computing, we first re-create two sub datasets from this dataset as shown in Table \ref{tab:datasets_details}. The first one is denoted as \textbf{PVAG} which consists of 8 plant disease classes from apple and grape. PVAG is a relatively large-scale dataset with balanced data distribution. The second is made up of 4 plant disease classes of images from peach and strawberry. This dataset is abbreviated as \textbf{PVPS} with unbalanced data structure. 

In addition, we utilise another public annotated plant disease image dataset, \textbf{Citrus Disease Dataset (CDD)} \citep{rauf2019citrus}, to evaluate the clustering algorithm, and the details of CDD are shown in Table \ref{tab:datasets_details}. CDD is a small-scale dataset with an unbalanced data structure. Given the above three datasets (the re-created two and CDD), we are able to cover cases for balanced and unbalanced data. 

\begin{table}
    \centering
    \caption{Details of Datasets. }
    \begin{center}
\footnotesize
 \begin{tabular}{||p{0.12\textwidth}|p{0.12\textwidth}|p{0.4\textwidth}|p{0.12\textwidth}||} 
 \hline
 Dataset & Plant & Disease / Health & Image Amount \\ [0.5ex] 
 \hline\hline
 \multirow{8}{*}{PVAG} & \multirow{4}{*}{Apple} & Scrab & 630 \\ 
 && Black Rot & 621 \\
 && Cedar Rust & 275 \\
 && Healthy & 1645  \\
 \cline{2-4}
 &\multirow{4}{*}{Grape} & Black Rot & 1180 \\
 && Black Measles (Esca) & 1383 \\
 && Healthy & 423 \\
 && Leaf Blight & 1076 \\
 \hline
 \multirow{4}{*}{PVPS} & \multirow{2}{*}{Peach} 
 & Bacterial Spot & 2298 \\ 
 && Healthy & 361 \\
 \cline{2-4}
 &\multirow{2}{*}{Strawberry} & Healthy & 457 \\ 
 && Leaf Scorch & 1110 \\
 \hline
 \multirow{5}{*}{CDD} & \multirow{5}{*}{Citrus} 
 & Black Spot & 172 \\ 
 && Canker & 164 \\
 && Greening & 205 \\
 && Healthy & 59 \\
 && Melanose & 14 \\
 \hline
\end{tabular}
\end{center} 
    \label{tab:datasets_details}
\end{table}
% The proposed clustering framework does not require any assumption or prior knowledge on the data structure of the datasets.

\begin{figure*}[ht]
\begin{subfigure}{.33\textwidth}
  \centering
  % include first image
  \includegraphics[width=.95\linewidth]{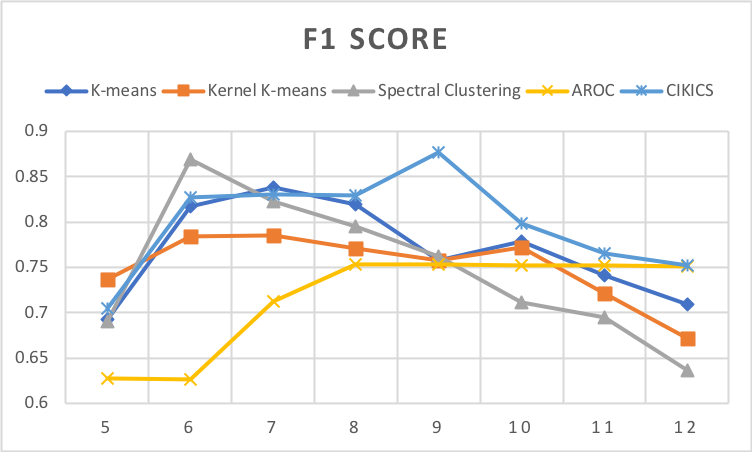}  
  \caption{}
  \label{fig:pvag_comparison_f1}
\end{subfigure}
\begin{subfigure}{.33\textwidth}
  \centering
  % include second image
  \includegraphics[width=.95\linewidth]{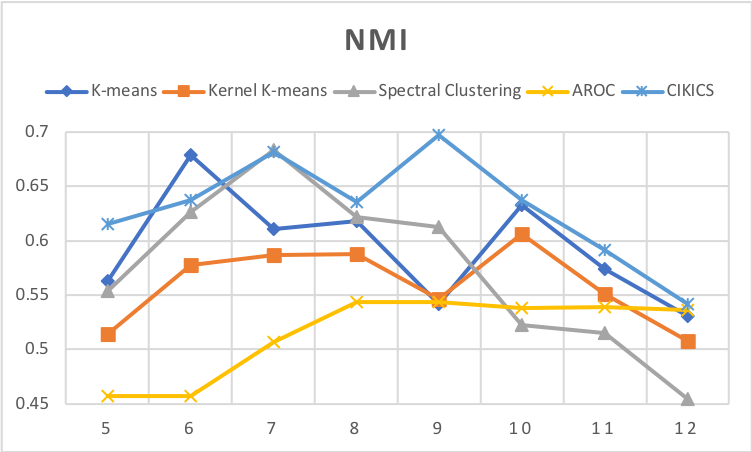}  
  \caption{}
  \label{fig:pvag_comparison_nmi}
\end{subfigure}
\begin{subfigure}{.33\textwidth}
  \centering
  % include second image
  \includegraphics[width=.95\linewidth]{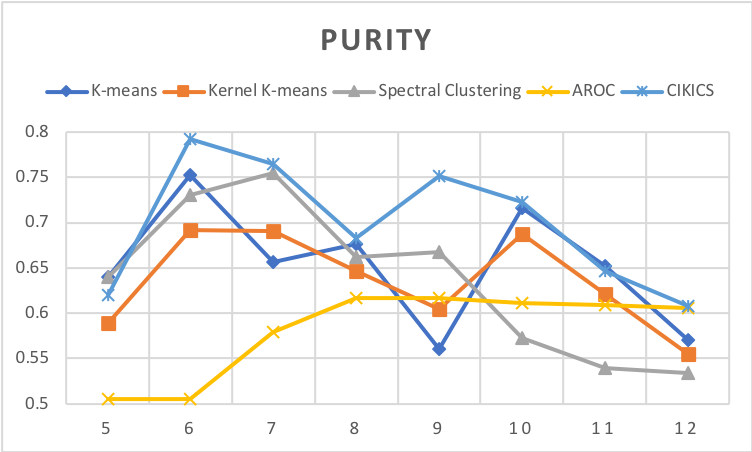}  
  \caption{}
  \label{fig:pvag_comparison_pur}
\end{subfigure}
\caption{Evaluation results with different pre-defined cluster numbers on PVAG. }
\label{fig:pvag_comparison}
\end{figure*}

\subsection{Evaluation Metrics}

We use external pairwise evaluation metrics \citep{otto2017clustering} to calculate $Precision$, $Recall$ and $F1$ score. We also employ the purity ($PUR$) and normalised mutual information ($NMI$) \citep{tian2017deepcluster} as evaluation metrics. 

Since above metrics can only reveal the overall quality of the clustering results, we develop an additional accuracy evaluation metric on the quality of each individual cluster. Since we aim to address the annotation of an unlabeled dataset, the high accuracy of each cluster is also crucial in evaluation. Therefore, we design a Per-cluster Quality Measure. For simplicity, we calculate the percentages of all same-label elements in each cluster, and find the highest one as the cluster accuracy which is denoted as $Acc$. We can also leverage $Acc$ to demonstrate the overall performance. The average accuracy of all clusters ($\displaystyle Ave Acc = \frac{\sum_{i=1}^{h}Acc_{i}}{h}$), where $h$ is the number of clusters, and the lowest accuracy amongst all clusters ($Min Acc = \min (Acc_{i})$) can be calculated. Also, in the resulting clusters, the percentage of \(Acc\) that is 100\% is denoted as $\displaystyle FPP=\frac{\sum_{i=1}^{h}1[Acc_{i}=1]}{h}$. And the percentage of \(Acc\) that is lower than 80\% is denoted as $\displaystyle LPP=\frac{\sum_{i=1}^{h}1[Acc_{i}<0.8]}{h}$. 

\subsection{Baselines}
We compare our CIKICS to existing clustering methods (including both classic and recent ones). We select 5 baselines which are divided into 2 different types of clustering algorithms: (1) algorithms without requiring the pre-defined number of clusters (including those that require hyper-parameter/threshold settings) and (2) algorithms with requiring the number of clusters as input. The first category of algorithms are AROC \citep{otto2017clustering} and FINCH \citep{sarfraz2019efficient}, where AROC takes thresholds as input, and FINCH requires no parameters at all. Both of them are recently proposed. The second type of algorithms include K-means \citep{arthur2006k}, Kernel K-means \citep{dhillon2004unified} and Spectral clustering \citep{ng2002spectral}.

\subsection{Results}

\subsubsection{\textbf{Comparison on balanced dataset}}
In Table \ref{tab:precision_table}, we compare the performance of our CIKICS with the 5 clustering methods on the balanced dataset, PVAG.

\begin{table}[ht]
    \centering
    \caption{Clustering performance on PVAG (\textit{pt.1})}
    \begin{center}
\begin{scriptsize}
 \begin{tabular}{||p{0.14\textwidth}|p{0.08\textwidth}|p{0.08\textwidth}|p{0.08\textwidth}|p{0.08\textwidth}|p{0.08\textwidth}|p{0.08\textwidth}||} 
 \hline
 Algorithm & Clusters & PRE & REC & F1 & NMI & PUR \\ [0.5ex] 
 \hline\hline
 K-means \citep{arthur2006k} & 8 & 0.878 & 0.770 & 0.820 & 0.618 & 0.676  \\
 \hline
 KKM \citep{dhillon2004kernel} & 8 & 0.887 & 0.776  & 0.828 & 0.564 & 0.638 \\ 
 \hline
 Spectral \citep{ng2002spectral} & 8 & 0.789 & 0.800 & 0.795 & 0.622 & 0.662  \\
 \hline
 AROC (0.4) \citep{otto2017clustering} & 8 & 0.605 & \textbf{0.995} & 0.753 & 0.544 & 0.617 \\ 
 \hline
 \multirow{5}{0.07\textwidth}{FINCH \citep{sarfraz2019efficient} } 
  & 2222 & 0.831 & 0.002 & 0.004 & 0.003 & 0.231 \\
  & 707 & 0.822 & 0.008 & 0.016 & 0.020 & 0.235 \\
  & 197 & 0.821 & 0.031 & 0.060 & 0.047 & 0.266 \\
  & 49 & 0.808 & 0.128 & 0.222 & 0.183 & 0.310 \\
  & 22 & 0.790 & 0.373 & 0.507 & 0.375 & 0.472 \\
 \hline
%  KIBICS-C & 16 & \textbf{0.965} & 0.373 & 0.539 & 0.457 & 0.484 \\
%  \hline
%  KIBICS-S & 30 & 0.970 & 0.232 & 0.374 & 0.363 & 0.459 \\
%  \hline
 CIKICS & 8 & \textbf{0.895} & 0.797 & \textbf{0.835} & \textbf{0.636} & \textbf{0.683}\\
 \hline
%  KIBICS-A & 8 & 0.968 & 0.544 & 0.697 & 0.783 & 0.573\\
%  \hline
%  KIBICS & 8 & 0.968 & 0.544 & 0.697 & 0.783 & 0.573\\
%  \hline
\end{tabular}
\end{scriptsize}
\end{center}
    \label{tab:precision_table}
\end{table}

The proposed method (last row) generates the highest clustering Precision, F1 score, NMI and Purity. On the other hand, AROC has a highest recall value on PVAG. However, if we look into further the data distribution in the generated clusters in Table \ref{tab:data_distribution_pvag}, apparently, AROC was particularly developed to cluster data with an unbalanced structure. Since precision and recall metrics involve errors \cite{otto2017clustering} \footnote{In clustering, each sample is taken as an individual cluster, it will have a higher accuracy but a lower recall rate. On the contrary, all samples are clustered in the same cluster will have a higher F1 and recall rate, but a low precision.}, AROC clustering produces the highest recall. 

From Table \ref{tab:data_distribution_pvag}, K-means, Kernel K-means, Spectral clustering and CIKICS can push data balancedly into clusters. This leads them to comparable performance if they are parameterised similarly. By contrast, the data distribution of CIKICS is closer to the ground-truth data structure (Table \ref{tab:datasets_details}. On the other hand, AROC has been proved with the capability of dealing with well unbalanced and large-scale data. It clusters data based on the pre-defined threshold. We test several thresholds to show how AROC performs on PVAG, which reveals the limitation and inflexibility of AROC on the balanced data. FINCH was proposed with the gimmick of a parameter-free clustering algorithm, but its performance on PVAG is less satisfactory, and we simply exclude it in further comparisons.

\begin{table*}[ht]
    \centering
    \caption{Numbers of images in each cluster for compared methods. }
    \begin{center}
\begin{scriptsize}
 \begin{tabular}{||p{0.15\textwidth}|p{0.05\textwidth}|p{0.05\textwidth}|p{0.05\textwidth}|p{0.05\textwidth}|p{0.05\textwidth}|p{0.05\textwidth}|p{0.05\textwidth}|p{0.05\textwidth}||} 
 \hline
 Cluster \# & 1 & 2 & 3 & 4 & 5 & 6 & 7 & 8 \\ [0.5ex] 
 \hline\hline
 K-means \citep{arthur2006k} & 751 & 1128 & 1075 & 640 & 614 & 1436 & 731 & 857 \\
 \hline
 KMM \citep{dhillon2004kernel} & 954 & 1398 & 786 & 548 & 1075 & 755 & 654 & 1062 \\
 \hline
 Spectral \citep{ng2002spectral} & 931 & 1075 & 894 & 964 & 1090 & 1008 & 690 & 580 \\
 \hline
 AROC \citep{otto2017clustering} & 631 & 2564 & 2260 & 279 & 421 & 1073 & 2 & 2 \\
 \hline
 CIKICS \citep{sarfraz2019efficient} & 697 & 1193 & 1383 & 1047 & 1089 & 1086 & 423 & 314 \\
 \hline
 \end{tabular}
\end{scriptsize}
\end{center} 
    \label{tab:data_distribution_pvag}
\end{table*}

\begingroup
\begin{figure*}[ht]
\begin{subfigure}{.33\textwidth}
  \centering
  % include first image
  \includegraphics[width=.95\linewidth]{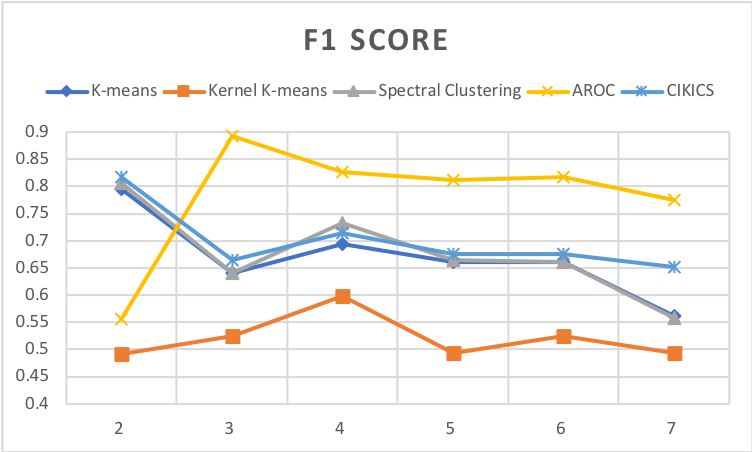}  
  \caption{}
  \label{fig:pvps_comparison_f1}
\end{subfigure}
\begin{subfigure}{.33\textwidth}
  \centering
  % include second image
  \includegraphics[width=.95\linewidth]{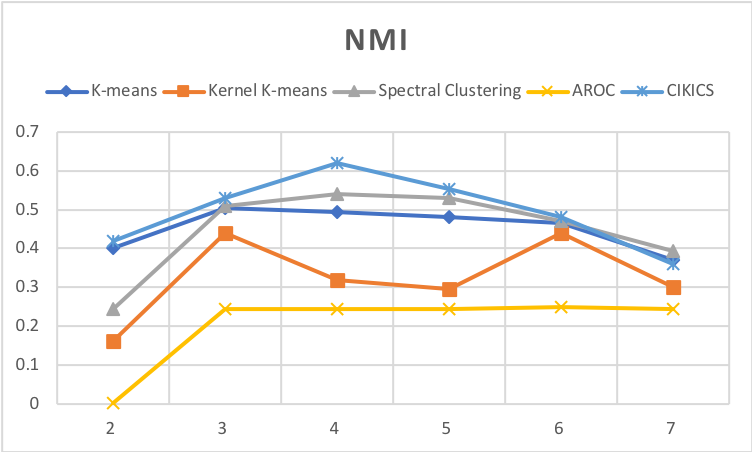}  
  \caption{}
  \label{fig:pvps_comparison_nmi}
\end{subfigure}
\begin{subfigure}{.33\textwidth}
  \centering
  % include second image
  \includegraphics[width=.95\linewidth]{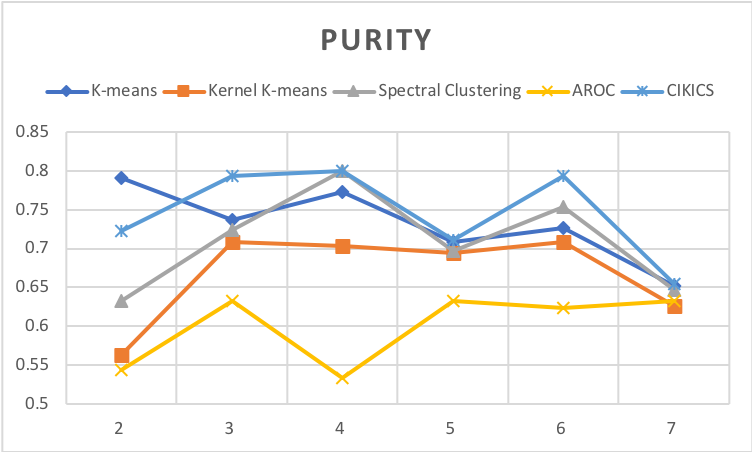}  
  \caption{}
  \label{fig:pvps_comparison_pur}
\end{subfigure}
\caption{Evaluation results with different pre-defined cluster numbers on PVPS. }
\label{fig:pvps_comparison}
\end{figure*}
\endgroup

 \begin{figure*}[ht]
\begin{subfigure}{.33\textwidth}
  \centering
  % include first image
  \includegraphics[width=.95\linewidth]{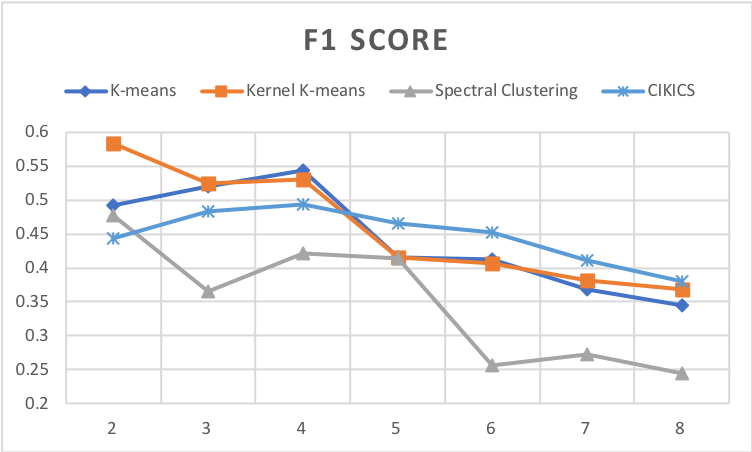}  
  \caption{}
  \label{fig:cdd_comparison_f1}
\end{subfigure}
\begin{subfigure}{.33\textwidth}
  \centering
  % include second image
  \includegraphics[width=.95\linewidth]{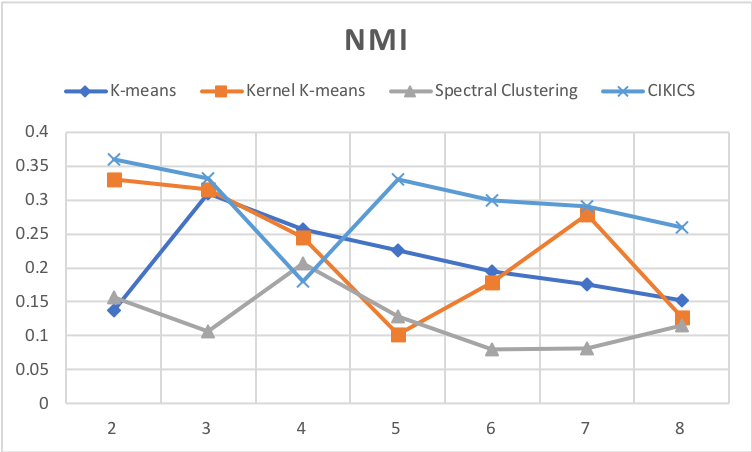}  
  \caption{}
  \label{fig:cdd_comparison_nmi}
\end{subfigure}
\begin{subfigure}{.33\textwidth}
  \centering
  % include second image
  \includegraphics[width=.95\linewidth]{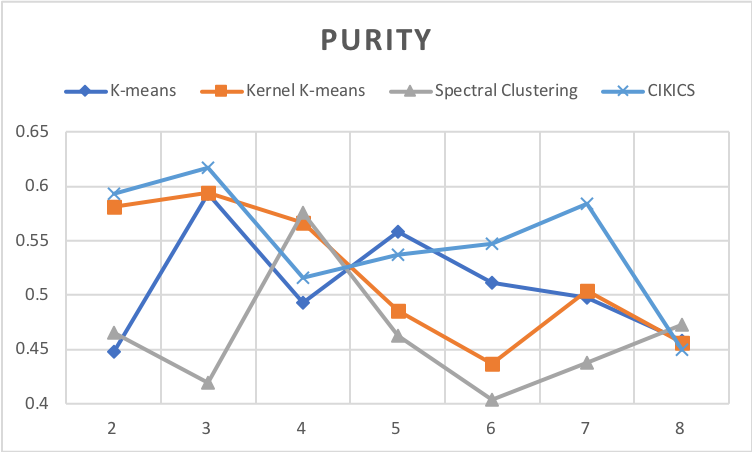}  
  \caption{}
  \label{fig:cdd_comparison_pur}
\end{subfigure}
\caption{Evaluation results with different pre-defined cluster numbers on CDD. }
\label{fig:cdd_comparison}
\end{figure*}

\begin{table}[t!]
    \centering
    \caption{Clustering performance on PVAG (\textit{pt.2}). }
    \begin{center}
\begin{scriptsize}
 \begin{tabular}{||p{0.2\textwidth}|p{0.12\textwidth}|p{0.12\textwidth}|p{0.12\textwidth}|p{0.12\textwidth}||} 
 \hline
 Algorithm & Ave Acc & Min Acc & FPP & LPP \\ [0.5ex] 
 \hline\hline
 K-means \citep{arthur2006k} & 0.878 & 0.553 & 0.125 & 0.250\\
 \hline
 KKM \citep{dhillon2004kernel} & 0.831 & 0.544 & 0.125 & 0.375 \\
 \hline
 Spectral \citep{ng2002spectral} & 0.868 & 0.558 & 0.125 & 0.250\\
 \hline
  FINCH (22) \citep{otto2017clustering} & 0.881 & 0.550 & 0.020 & 0.163\\
 \hline
 AROC (0.4) \citep{sarfraz2019efficient} & 0.904 & 0.539 & 0.500 & 0.250\\
 \hline
 CIKICS & \textbf{0.891} & \textbf{0.556} & \textbf{0.125} & \textbf{0.125} \\
 \hline
 \end{tabular}
\end{scriptsize}
\end{center} 
    \label{tab:ave_acc_pvag}
\end{table}

As for the per-cluster accuracy, we list them in Table \ref{tab:ave_acc_pvag}. Interestingly, our proposed framework outperforms other methods. Figure \ref{fig:pvag_comparison} intuitively demonstrates that CIKICS has the outstanding performance with different cluster numbers (around the ground-truth class number). For K-means, Kernel K-means, Spectral clustering and CIKICS, the clustering performance is generally better when the cluster number is closer to the real number of classes. Nevertheless, the performance of AROC changes gently, except the increase when the cluster number approaches the ground-truth number of classes. Furthermore, we can tell that sometimes other clustering methods outperform ours in these line plots. For example, as shown in Figure \ref{fig:pvag_comparison_f1}, the F1 score of spectral clustering at 6 clusters is slightly higher than others because it divides the smallest sized class in the ground-truth 8 classes balancedly into other 5 clusters. At the same time, as shown in Figure \ref{fig:pvag_comparison_nmi}, K-means demonstrates higher NMI measure at 6 clusters, which means the clustering result is highly correlated with the ground-truth. As known, there are 8 classes in the dataset. When we look into the result we found that images in two classes with the smallest size have been separated evenly into the other 6 clusters. This enables the other 6 classes dominate the resulted 6 clusters, which increases the mutual information between the result and the ground truth.

\subsubsection{\textbf{Comparison on unbalanced datasets}}
To further evaluate the capabilities of dealing with unbalanced data, we use the re-created PVPS from PlantVillage. The performance is shown in Table \ref{tab:pvps_precision_table}.

\begin{table}[t!]
    \centering
    \caption{Clustering performance on PVPS (\textit{pt.1}). }
    \begin{center}
\begin{scriptsize}
 \begin{tabular}{||p{0.14\textwidth}|p{0.08\textwidth}|p{0.08\textwidth}|p{0.08\textwidth}|p{0.08\textwidth}|p{0.08\textwidth}|p{0.08\textwidth}||} 
 \hline
 Algorithm & Clusters & PRE & Rec & F1 & NMI & PUR \\ [0.5ex] 
 \hline\hline
 K-means \citep{arthur2006k} & 4 & 0.835 & 0.594 & 0.694 & 0.493 & 0.773
  \\
 \hline
 KKM \citep{dhillon2004kernel} & 4 & 0.658 & 0.547 & 0.598 & 0.319 & 0.703
 \\ 
 \hline
 Spectral \citep{ng2002spectral} & 4 & 0.841 & 0.602 & 0.702 & 0.541 & 0.800
 \\
 \hline
 AROC (0.05) \citep{otto2017clustering} & 4 & 0.705 & \textbf{0.999} & \textbf{0.827} & 0.245 & 0.533
  \\
 \hline
 \multirow{4}{0.075\textwidth}{FINCH \citep{sarfraz2019efficient}} 
  & 1367 & 0.923 & 0.001 & 0.003 & 0.004 & 0.544 \\
  & 430 & 0.928 & 0.006 & 0.012 & 0.013 & 0.544 \\
  & 125 & 0.926 & 0.025 & 0.049 & 0.033 & 0.544 \\
  & 69 & 0.931 & 0.067 & 0.124 & 0.145 & 0.550 \\
 \hline
%  KIBICS-C & 30 & 0.989 & 0.043 & 0.083 & 0.213 & 0.307 \\
%  \hline
%  KIBICS-S & 15 & \textbf{0.994} & 0.104 & 0.188 & 0.313 & 0.459 \\
%  \hline
 CIKICS & 4 & \textbf{0.849} & 0.657 & 0.715 & \textbf{0.619} & \textbf{0.800} \\
 \hline
\end{tabular}
\end{scriptsize}
\end{center} 
    \label{tab:pvps_precision_table}
\end{table}

\begin{table}[t!]
    \centering
    \caption{Clustering performance on PVPS (\textit{pt.2}). }
    \begin{center}
\begin{scriptsize}
 \begin{tabular}{||p{0.2\textwidth}|p{0.12\textwidth}|p{0.12\textwidth}|p{0.12\textwidth}|p{0.12\textwidth}||} 
 \hline
 Algorithm & Ave Acc & Min Acc & FPP & LPP \\ [0.5ex] 
 \hline\hline
 K-means \citep{arthur2006k} & 0.888 & 0.711 & 0.250 & 0.250\\
 \hline
 KKM \citep{dhillon2004kernel} & 0.891 & 0.598 & 0.250 & 0.250\\
 \hline
 Spectral \citep{ng2002spectral} & 0.901 & 0.715 & 0.250 & 0.250\\
 \hline
 FINCH (69) \citep{otto2017clustering} & 0.917 & 0.471 & 0.455 & 0.182\\
 \hline
 AROC (0.05) \citep{sarfraz2019efficient} & 0.920 & 0.333 & \textbf{0.803} & \textbf{0.175}\\
 \hline
 CIKICS & \textbf{0.928} & \textbf{0.715} & 0.000 & 0.250 \\
 \hline
 \end{tabular}
\end{scriptsize}
\end{center} 
    \label{tab:ave_acc_pvps}
\end{table}

AROC is generally better in F1 score according to Table \ref{tab:pvps_precision_table} and Figure \ref{fig:pvps_comparison}. However, as shown in Table \ref{tab:ave_acc_pvps}, the \textit{Min Acc} is the lowest, which indicates there is still room for improvement on AROC. Its \textit{Ave Acc} arrives at 0.928. Impressively, its \textit{FPP} reaches 0.803, i.e. \(4/5\) clusters has 100\% same-class identities. Also, the clustering result on PVPS produced by AROC induces the lowest \textit{LPP}. %These demonstrate the overwhelming advantage of AROC over the unbalanced data. %Despite that the size of PVPS may not be large enough to represent itself as a large scale dataset, PVPS is large enough to guarantee the great performance of AROC. 
Interestingly, our framework is still comparable to AROC over the dataset with an unbalanced structure with 0.619 NMI, 0.8 purity, 0.928 \textit{Ave Acc} and 0.715 \textit{Min Acc}. We observe that the proposed method has a noticeable applicability and scalability for both unbalanced and balanced data, especially when comparing it with K-means, Kernel K-means and Spectral clustering. 

Besides the above comparisons, the comparative results on small-scale balanced data is shown in Table \ref{tab:cdd_precision_table} and \ref{tab:ave_acc_cdd}. AROC has been removed from the comparative study on CDD, since we observe through experiments that AROC has a great limitation on handling small-scale data. Specifically, AROC relies on the pre-defined threshold, and we test the threshold from $1000$ to $0.001$ (with a step size of $50$ from $1000$ to $100$, $10$ from $100$ to $10$, $1$ from from $10$ to $1$, $0.1$ from $1$ to $0.1$, then $0.01$ and $0.001$). Unfortunately, we always obtain a single cluster using AROC. This demonstrates that our method is more robust than AROC to small-scale datasets. %Again, this confirms the feasibility and transferability of CIKICS. 

\begin{table}[t!]
    \centering
    \caption{Clustering performance on CDD (\textit{pt.1}). }
    \begin{center}
\begin{scriptsize}
 \begin{tabular}{||p{0.13\textwidth}|p{0.08\textwidth}|p{0.08\textwidth}|p{0.08\textwidth}|p{0.08\textwidth}|p{0.08\textwidth}|p{0.08\textwidth}||} 
 \hline
 Algorithm & Clusters & PRE & Rec & F1 & NMI & PUR \\ [0.5ex] 
 \hline\hline
 K-means \citep{arthur2006k} & 5 & 0.487 & 0.362 & 0.415 & 0.226 & 0.558 \\
 \hline
 KKM \citep{dhillon2004kernel} & 5 & 0.468 & 0.373 & 0.415 & 0.102 & 0.486 \\ 
 \hline
 Spectral \citep{ng2002spectral} & 5 & 0.486 & 0.360 & 0.414 & 0.129 & 0.463 \\
 \hline
 \multirow{3}{0.07\textwidth}{FINCH \citep{sarfraz2019efficient}} 
  & 201 & 0.599 & 0.009 & 0.018 & 0.021 & 0.353 \\
  & 62 & 0.576 & 0.036 & 0.068 & 0.081 & 0.425 \\
  & 20 & 0.551 & 0.127 & 0.206 & 0.071 & 0.412 \\
 \hline
%  KIBICS-C & 5 & 24 & \textbf{0.720} & 0.094 & 0.166 & 0.143 & 0.438 \\
%  \hline
%  KIBICS-S & 5 & 13 & 0.714 & 0.216 & 0.332 & \textbf{0.444} & \textbf{0.544} \\
%  \hline
 CIKICS & 5 & \textbf{0.521} & \textbf{0.422} & \textbf{0.466} & \textbf{0.331} & \textbf{0.537} \\
 \hline
\end{tabular}
\end{scriptsize}
\end{center} 
    \label{tab:cdd_precision_table}
\end{table}

\begin{table}[t!]
    \centering
    \caption{Clustering performance on CDD (\textit{pt.2}). }
    \begin{center}
\begin{scriptsize}
 \begin{tabular}{||p{0.2\textwidth}|p{0.12\textwidth}|p{0.12\textwidth}|p{0.12\textwidth}|p{0.12\textwidth}||} 
 \hline
 Algorithm & Ave Acc & Min Acc & FPP & LPP \\ [0.5ex] 
 \hline\hline
 K-means \citep{arthur2006k} & 0.670 & 0.571 & 0.000 & 0.800\\
 \hline
 KKM \citep{dhillon2004kernel} & 0.778 & 0.514 & 0.400 & 0.600\\
 \hline
 Spectral \citep{ng2002spectral} & 0.736 & 0.593 & 0.000 & 0.800\\
 \hline
 FINCH (20) \citep{sarfraz2019efficient} & 0.685 & 0.366 & 0.000 & 0.750\\
 \hline
 CIKICS & \textbf{0.779} & \textbf{0.538} & \textbf{0.400} & \textbf{0.600} \\
 \hline
 \end{tabular}
\end{scriptsize}
\end{center} 
    \label{tab:ave_acc_cdd}
\end{table}
% https://www.rbtechblog.com/blog/cluster_accuracy

Regarding the clustering accuracy, CIKICS reaches 0.830 F1 score, 0.636 NMI, 0.683 purity and 0.891 \textit{Ave Acc} on PVAG (7323 images, 8 plant disease classes) in Table \ref{tab:precision_table} and \ref{tab:ave_acc_pvag}, which obviously outperforms other clustering algorithms. CIKICS achieves 0.715 F1 score, 0.616 NMI, 0.800 purity and 0.928 \textit{Ave Acc} on PVPS (4222 images, 4 plant disease classes) in Table \ref{tab:pvps_precision_table} and \ref{tab:ave_acc_pvps}. This is relatively inferior to AROC but still reveals the feasibility of CIKICS on unbalanced data. For the evaluation of clustering algorithms on CDD (604 images, 5 plant disease classes), CIKICS has shown the capability of dealing with the small-scale datasets with 0.466 F1 score, 0.331 NMI, 0.537 purity and 0.779 \textit{Ave Acc} in Table \ref{tab:cdd_precision_table} and \ref{tab:ave_acc_cdd}. The performance is significantly higher than the other clustering algorithms on CDD. Moreover, according to Figure \ref{fig:cdd_comparison}, we can see that different pre-defined cluster numbers lead to significant up-and-downs on F1 score, NMI and purity measures. Specifically, the outstanding part appears generally for 4 or 5 clusters. This is mainly due to the challenge of the small scale of the dataset. As the ground-truth class number of CDD is 5, there is an obvious limitation for CIKICS to produce highly accurate clusters as training set to learn deep models for further image classification. When the pre-defined cluster number is more than 5, we can tell that the performance of CIKICS is generally better than others because the larger pre-defined cluster number provides more potential for CIKICS to cluster every image into the correct group.

\subsubsection{\textbf{Evaluation on deep learning module}}

To evaluate the performance with part or the whole deep learning module in CIKICS, we conduct an ablation study in Table \ref{tab:learning_comparison}. It is obvious that CIKICS involving all three deep learning models outperforms other combinations. The image classifier is always included, as it is fast to produce a potential class list for each non-clustered image. %Both the similarity finder and the Siamese network leverage pairwise similarity to recognise images, which is time-consuming. This is also why we construct the framework to increase the accuracy, at the same time, to restrain the runtime.

\begin{table}[t!]
    \centering
    \caption{Evaluation of deep learning module on PVAG. IM: image classifier. SF: pre-trained CNN-based similarity finder. SN: the Siamese network.}
    \begin{center}
\begin{scriptsize}
 \begin{tabular}{||p{0.14\textwidth}|p{0.08\textwidth}|p{0.08\textwidth}|p{0.08\textwidth}|p{0.08\textwidth}|p{0.08\textwidth}|p{0.08\textwidth}||} 
 \hline
 Models & Clusters & PRE & Rec & F1 & NMI & PUR \\ [0.5ex] 
 \hline\hline
 IM & 8 & 0.883 & 0.736 & 0.803 & 0.632 & 0.675 \\
 \hline
 IM\&SF & 8 & 0.865 & 0.751 & 0.804 & 0.606 & 0.643 \\ 
 \hline
 IM\&SN & 8 & 0.887 & 0.735 & 0.804 & 0.635 & 0.680 \\
 \hline
 CIKICS & 8 & \textbf{0.895} & \textbf{0.797} & \textbf{0.835} & \textbf{0.636} & \textbf{0.683}\\
 \hline
\end{tabular}
\end{scriptsize}
\end{center} 
    \label{tab:learning_comparison}
\end{table}

%From the previous comparative experiments, we find an interesting fact that AROC always keeps the high recall, and consequently achieves high F1 score. However, when we look into the details of the resulted clusters, they are obviously not satisfactory enough to support the outstanding evaluation performance. Especially in the comparative study on CDD , we had already defined the threshold from 1000 to 0.001, but what we had obtained was 1 cluster containing all images as the clustering result with both 1.000 recall, and 0.426 and 0.494 F1 score respectively, which is apparently not desired and satisfying. Therefore, the importance of an as exhaustive as possible performance assessment has been reflected, which is why we involved NMI, purity and our proposed per accuracy measures in the performance evaluation. 

%The result indicates strengths and weaknesses in our proposed clustering framework. 
\textbf{Remark.}
CIKICS is devised on top of Kernel K-means which is generally used to process balanced dataset. With integrating image classification and similarity measurements in our framework, CIKICS has the ability to group the similar elements without considering the ground-truth data structure. It is also robust to different scales of datasets. In brief, the proposed CIKICS has been provided with a sound capability in handling unbalanced/balanced data and large/small-scale data. 
% However, for the weakness of CIKICS, two deep learning techniques in CIKICS, especially image classification, highly rely on a large number of "ground-truth" training samples. As a result, the clustering part of CIKICS is highly expected to produce clusters that highly match the ground truth. 
%which could not be achieved 100\%, because (i) we could not dominate the size of the applied dataset, e.g. a small sized dataset with few elements in each class, and (ii) 

\section{Conclusion}
In this paper, we have presented a novel clustering framework, CIKICS, for grouping unlabelled plant disease images without any supervision. This method is built on our key observation that Kernel K-means can produce some accurate clusters and one chaotic cluster (i.e., abnormal cluster). This leads us to devise a cross iterative clustering algorithm to generate pseudo-labeled clusters and a chaotic cluster to be classified by further deep learning. It is flexible and robust to data organisation structure (balanced or unbalanced) and scales of datasets. Extensive experiments demonstrate that the proposed CIKICS has an outstanding performance comparing to existing clustering methods, in terms of quantitative evaluations. 

\bibliographystyle{unsrt}
\bibliography{references}

\end{document}